\documentclass[11pt,a4paper]{article}
\usepackage[hyperref]{naaclhlt2019}
\usepackage{times}
\usepackage{amsmath}
\usepackage{latexsym}
\usepackage{subcaption}
\usepackage{booktabs}
\usepackage[inline]{enumitem}
\usepackage{graphicx}

\aclfinalcopy 
\def\aclpaperid{733} 

\title{Resilient Combination of Complementary CNN and RNN Features for Text Classification through Attention and Ensembling}

\author{Athanasios Giannakopoulos$^{\dagger}$, Maxime Coriou$^{\dagger}$\thanks{\hspace{6pt}Work done during the master thesis of Maxime Coriou at Data, Analytics \& AI | Swisscom AG.}\protect\phantom{\footnotesize 1}, Andreea Hossmann, \\ 
\bf{Michael Baeriswyl} \and \bf{Claudiu Musat} \\
Data, Analytics \& AI Group | Swisscom AG \\ 
{\tt firstname.lastname@swisscom.com} \\
$^{\dagger}$\textit{Equal contribution}}

\date{}

\begin{document}
\maketitle

\begin{abstract}
State-of-the-art methods for text classification include several distinct steps of pre-processing, feature extraction and post-processing. In this work, we focus on end-to-end neural architectures and show that the best performance in text classification is obtained by combining information from different neural modules. Concretely, we combine convolution, recurrent and attention modules with ensemble methods and show that they are complementary. We introduce ECGA, an end-to-end go-to architecture for novel text classification tasks. We prove that it is efficient and robust, as it attains or surpasses the state-of-the-art on varied datasets, including both low and high data regimes.
\end{abstract}

\section{Introduction} \label{sec:intro}

Text classification is among the most common natural language processing problems. Its applications vary from separating documents into classes~\citep{yang2016hierarchical} to finding argumentative phrases~\citep{Fierro2017} or detecting churny tweets~\citep{amiri2015target}.
Techniques range from traditional \textit{tf-idf} methods to modern deep neural networks. Generally, traditional methods are used for simpler classification spaces and in low data regimes~\citep{amiri2015target}. However, neural modules are becoming the norm for complex problems with higher data availability. 

Modern text classifiers use mainly neural modules for feature extraction. Convolutional neural networks (CNN)~\citep{lecun1998gradient} or embedded convolution layers in larger networks can be used as feature extractors because of their location invariance property. Recurrent neural units such as Gated Recurrent Units (GRU)~\citep{cho2014learning} are used~\citep{socher2013recursive} because of their sequence modelling capabilities. Finally, the use of attention~\citep{li2017joint} is constantly growing since it can tackle the forgetfulness of recurrent cells on long sequences~\citep{D15-1166}.

Dropout~\citep{srivastava2014dropout} and ensemble methods (e.g. Random Forests (RF)~\citep{breiman2001random}) are two popular countermeasures for overfitting, which is a constant risk for deep learning models trained in low data regimes. In the case of ensemble methods, predictions of multiple learners are combined to form the final prediction.

In this work, we combine all aforementioned components and introduce a widely applicable text classifier: an Ensemble of CNN-GRU-Attention, hereafter denoted as ECGA. ECGA benefits from the complementary feature representation capacities of the three neural modules it exploits. At the same time, it constitutes an efficient way of limiting overfitting since it is based on ensemble methods, i.e. the final prediction is done by averaging the predictions from multiple learners. We are aware that individual neural components are widely used in text classification either individually or in pairs of two. However, combining all of them in one text classifier is novel.
 
We deploy and test ECGA in three different text classification tasks, namely
\begin{enumerate*}[label=(\roman*)]
    \item argumentation mining,
    \item topic classification and 
    \item textual churn detection.
\end{enumerate*}
The first two tasks are complex multi-class classification problems with large datasets containing up to 44 classes. The third task is a binary classification, however the nature of the text and the task is difficult even for human annotators\footnote{Confirmed by the low annotation confidence in~\citep{amiri2015target}.}. The dataset for the third task is small, which forces ECGA to operate in low data regimes.

The first finding that emerges from our results is that ECGA \emph{exceeds or at least attains the state-of-the-art} in all aforementioned classification tasks. It does so in an end-to-end way without any changes to its architecture, except for hyper-parameter tuning. This resilience makes ECGA a prime choice for new tasks, as it even outperforms architectures that were tailored for the studied tasks.

The second finding is that everything matters, i.e. \emph{ensemble methods combined with all neural modules lead to a performance increase}. By gradually adding complementary neural components we obtain sustained performance  increases.


\section{Related Work} \label{sec:related work}

\citet{Fierro2017} have contributed the most on argumentation mining after releasing a dataset with more than $200000$ arguments. The best performance on this dataset is based on the FastText classifier~\citep{joulin2016bag}.

With respect to topic classification,~\citet{zhang2015character} created the DBpedia dataset for multi-class text classification. Numerous research teams (e.g.~\citet{johnson2016supervised} and~\citet{johnson2017deep}) have worked on DBpedia by applying different models and feature extraction methods. Lately,~\citet{howard2018fine} employed transfer learning and achieved the state-of-the-art on this dataset.

~\citet{amiri2016short} performed textual churn detection using tweets about 3 mobile providers and obtained their best results by using recurrent cells. Later,~\citet{gridach2017churn} improved the performance by adding hand-crafted features based on logic rules to a CNN.

\section{ECGA Architecture} \label{sec:model architecture}

ECGA orchestrates all types of feature extraction and text classification modules. We want to show that the techniques of 
\begin{enumerate*}[label=(\roman*)]
    \item convolution,
    \item recurrence, 
    \item attention and
    \item ensembles
\end{enumerate*}
are complementary. 

\textbf{1.} We employ CNNs that are great feature extractors for text classification~\citep{yin2017comparative}. We create an $n \times m$ input matrix -- $n$ is the number of words of the input text and $m$ equals to the number of features -- and apply convolution on it with $f$ filters of kernel size $k$. Each filter slides over $k$ words (i.e. $k$-grams) and creates a vector of size $n-k+1$. We concatenate the output of the $f$ filters without max pooling and create an $(n-k+1) \times f$ matrix. Hence, the $j^{th}$ row of this matrix is a feature of the $j^{th}$ $k$-gram of the input sentence.

\textbf{2.} We then feed the output of the CNN into a bidirectional GRU (BiGRU) i.e. the input size of the BiGRU network is $n-k+1$. The output vector of each state embeds information about the structure of the input text learned from the sequences of $k$-grams.

\textbf{3.} We incorporate and apply attention on the output states of the BiGRU network~\citep{li2017joint}. This allows us to construct a final feature vector $\boldsymbol\alpha$ of the input text using a weighted sum of all the output states of the BiGRU network. The final layer of ECGA passes $\boldsymbol\alpha$ through a \textit{softmax} activation for the text classification.

\begin{figure}[t]
    \centering
    \includegraphics[width=77mm]{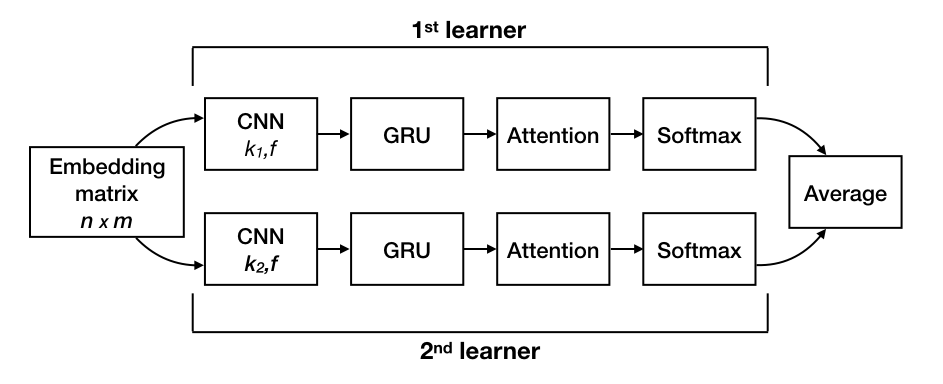}
    \caption{ECGA architecture with two learners.}
    \label{fig:ECGA}
\end{figure}

\textbf{4.} Finally, we exploit multiple learners, i.e. ensemble methods, in order to combine diverse predictions and attain higher performance. We do so by performing convolutions with different kernel sizes $k_{i}$ on the input matrix. This allows us to extract at the same time features for 2-grams, 3-grams, etc. by choosing different values for $k_{i}$. We then fork the deeper layers of the network (i.e. BiGRU, attention and \textit{softmax}) according to the number of different kernel sizes we use. In that way, we create multiple learners (similar to random forests) and train them using different features for the same task. The final prediction is done by averaging the predictions of all the learners. Figure \ref{fig:ECGA} shows ECGA with two learners.

\section{Experiments and Results} \label{sec:experiments-and-results}

We wish to prove that all techniques of 
\begin{enumerate*}[label=(\roman*)]
    \item convolutions,
    \item recurrent units,
    \item attention and
    \item ensembles
\end{enumerate*}
contribute in the performance increase. We use models that exploit only a subset of the available neural models as baselines and show that ECGA outperforms them, i.e. the best performance comes after combining ensemble methods with all available neural models. 

We chose three datasets that emphasize the diversity of situations that ECGA can perform in. We first aim for a large, well-studied dataset with a high number of classes. With its 14 classes, the DBpedia dataset overshadows others like \textit{AGNews}, that contains only 4.
We then focus on a different language (Spanish), in a classification setting with an even higher class count, concretely 44. Finally, we hypothesize that, despite its apparent size, ECGA can become the new state-of-the-art in a complex low data regime -- represented by the textual churn detection dataset~\citep{amiri2015target}. The complexity of the third task -- textual churn detection -- relies on two factors. First, the nature of the task is inherently difficult even for human annotators. Secondly, the available dataset is quite small and very unbalanced.

Finally, we did not focus on tasks where the state-of-the-art results are obtained mainly after heavy fine-tuning and pre-processing, a practice that does not generalize to new domains. Examples of this include sentiment analysis on datasets like \textit{IMDB}~\citep{maas-EtAl:2011:ACL-HLT2011} and \textit{YELP}.
ECGA achieves very good performance without any cumbersome data pre-processing. Moreover, we are not tackling a multi-label setting, therefore datasets like \textit{Reuters} are not suitable for our analysis.

In all experiments the hyper-parameter tuning consists of grid search, with at least 5 experiments for each setting. We do not report confidence intervals as conducting one experiment on the argumentation mining and DBpedia datasets takes more than $7h$ and $12h$ respectively. This is also the reason we do not perform experiments with more than two learners\footnote{Adding extra learners increases the model parameters, thus also the training time of the model.}.

\subsection{DBpedia}

The DBpedia dataset is compiled for multi-class text classification using Wikipedia article titles and abstracts. It contains datapoints from $14$ classes with a pre-defined train and test set~\citep{zhang2015character}.
The state-of-the-art performance on DBpedia is achieved by~\citet{howard2018fine} through a non end-to-end system that uses transfer learning. However, this implies the dataset availability from at least two similar domains and therefore we do not compare ECGA against their system. The best comparable method, which does not use transfer learning is that of \citet{johnson2016supervised}, who reach an error rate of $0.84\%$. 

For training, we use the FastText word embeddings~\citep{bojanowski2016enriching} and pad all sentences to a length of 60. For pure CNN, we use a kernel size of 2 with 256 filters. The number of units equals to 128 whenever the model contains GRU cells. ECGA has two learners with kernel sizes of 2 and 3. The number of filters is 256 and the number of units equals to 128 for both learners. We also apply dropout with a rate of 0.3 between all layers. Finally, we exploit the \textit{adam} optimizer with a learning rate of $10^{-4}$, $\beta_{1}=0.7$, $\beta_{2}= 0.99$. 

We present the model comparison on DBpedia in Table~\ref{tab:dbped}. ECGA beats all baselines and attains the state-of-the-art in DBpedia. In addition, it does so without the need of training or fine tuning word embeddings while being an end-to-end model.

\begin{table}[]
\centering
    \begin{tabular}{@{}lc@{}}
        \toprule
        & \textit{Error rate} (\%)\\\midrule
        ~\citet{johnson2016supervised} & 0.84 \\\midrule
        CNN & 1.29\\
        BiGRU+ATT & 0.88\\
        CNN+BiGRU & 0.87\\
        CNN+BiGRU+ATT & 0.85\\
        ECGA & \textbf{0.84} \\\bottomrule
    \end{tabular}
\caption{Performance comparison on DBpedia.}
\label{tab:dbped}
\end{table}

\subsection{Argumentation Mining}

We use the dataset released by~\citet{Fierro2017} for argumentation mining, our second complex text classification task, in Spanish instead of English. It contains more than $200000$ data points and each one is labelled with a topic, concept and argument mode. The dataset can be used for two different classification tasks (Task A and Task C~\citep{Fierro2017}) with up to $44$ labels.

To assure a fair comparison, we adopt exactly the experimental setup of~\citet{Fierro2017}. For Task A, we predict the concept of a given data point. To do so, we split the data points in four disjoint topic sets -- Values (V), Rights (R), Duties (D) and Institutions (I). We then train different classifiers on the four subsets in order to predict the concept. For Task C, we predict the argumentation mode of a data point after removing those with blank or undefined label. 

Once again, we use a padding of 60 tokens and the FastText word embeddings. With respect to the hyper-parameters, we exploit the \textit{adam} optimizer with its default parameter values\footnote{A learning rate of $10^{-3}$, $\beta_{1}=0.9$, $\beta_{2}= 0.999$}. For CNN, we use a kernel size of 2 with 256 (for topics V and R) or 512 (for topics D and I) filters. The number of units in the GRU layer equals to either 128 (for topics V and R) or 256 (for topics D and I). ECGA employs two learners with kernel sizes of 2 and 3. Both learners have 256 filters and 128 units independently of the topic. We apply dropout between all layers with a rate of 0.5. For Task C we use the same parameters as for Task A. The only difference is that ECGA uses two learners with kernel sizes of 2 and 3 with 512 filters and 256 units.

Experimental results for Task A and C are tabulated in Tables~\ref{tab:argumentation_A} and~\ref{tab:argumentation_C} respectively with the same layout as in~\citep{Fierro2017}. Once again, the performance we attain on both tasks proves that ECGA surpasses significantly all baselines and the state-of-the-art.

\begin{table}
    \begin{subtable}{0.5\textwidth}
        \centering
        \begin{tabular}{@{}lccccc@{}}
            \toprule
             & \multicolumn{5}{c}{Task A (\textit{Accuracy} \%)}\\ \midrule
             & V & R & D & I & Avg.\\\midrule
             FastText & 68.0 & 71.1 & 76.9 & 69.4 & 71.4 \\\midrule
             CNN & 70.0 & 72.0 & 76.0 & 70.4 & 72.1 \\
             BiGRU+ATT & 72.3 & 74.4 & 77.3 & 71.9 & 73.9\\
             CNN+BiGRU & 70.9 & 73.8 & 77.0 & 71.2 & 73.2\\
             CGA & 72.2 & 74.2 & 77.6 & 72.1 & 74.0\\
             ECGA & \textbf{72.5} & \textbf{75.0} & \textbf{78.2} & \textbf{72.4} & \textbf{74.5}\\\bottomrule
        \end{tabular}
        \caption{Argumentation mining: Task A. CGA stands for CNN + BiGRU + ATT.}
        \label{tab:argumentation_A}
        \vspace{5mm}
    \end{subtable}
    \begin{subtable}{0.5\textwidth}
        \centering
        \begin{tabular}{@{}lc@{}}
            \toprule
             & Task C (\textit{F-score} \%)\\ \midrule
             FastText & 65.4 \\\midrule
             CNN & 67.0\\
             BiGRU+ATT & 70.9\\
             CNN+BiGRU & 70.4\\
             CNN+BiGRU+ATT & 71.3\\
             ECGA & \textbf{71.6} \\\bottomrule
        \end{tabular}
        \caption{Argumentation mining: Task C.}
        \label{tab:argumentation_C}
    \end{subtable}
    \caption{Model performance on argumentation mining. The FastText baseline is from~\citet{Fierro2017}.}
    \label{tab:argumentation_result}
\end{table}

\subsection{Textual Churn Detection}

We use the publicly available dataset of~\citet{amiri2015target} for textual churn detection. The authors use only tweets with annotation confidence larger than $0.7$. We follow the same approach in order to have a fair comparison against their system. The resulting dataset contains 4728 tweets and only 900 out of them are churny.

\citet{gridach2017churn} achieve state-of-the-art in textual churn detection by enriching the features extracted from a CNN with hand-crafted ones. This approach does not scale, as additional human knowledge is not readily available in all cases.

We use the Twitter GloVe word embeddings and perform some data cleaning as standardization of URLs, smileys, usernames and numbers. In addition, we restrict our vocabulary to 1000 tokens and pad each tweet to a length of 50. We evaluate our models by performing 10-fold cross validation, same as~\citet{amiri2016short} and~\citet{gridach2017churn}.
The \textit{adam} optimizer has again the default parameter values. For CNN, we use a kernel size of 3 with 64 filters and 64 units for BiGRU. The kernel size equals to 2, the filters to 128 and the units to 64 when CNN is combined with BiGRU (with or without Attention). Finally, ECGA has two learners with kernel sizes of 1 and 2 with 128 filters and 64 units. 

The results of Table~\ref{tab:churn table} show once again that the more neural modules we add, the more the performance increases. ECGA surpasses the state-of-the-art in textual churn detection by $3.15\%$.

\begin{table}
    \centering
    \begin{tabular}{@{}lc@{}}
        \toprule
         & Macro \textit{F-score} (\%) \\\midrule
         \citet{gridach2017churn} & 83.85\\\midrule
         CNN & 81.94\\
         BiGRU+ATT & 84.21\\
         CNN+BiGRU & 84.48\\
         CNN+BiGRU+ATT & 86.26\\
         ECGA & \textbf{87.00}\\\bottomrule
        \end{tabular}
    \caption{Model performance on churn detection.}
    \label{tab:churn table}
\end{table}

\section{Conclusion} \label{sec:conclusion}

We work towards creating a one-size-fits-all go-to model for any novel text classification task. Our effort originates from our belief that all neural components can gradually contribute in the performance increase of a classifier. We introduce ECGA, a universal text classifier, that combines Ensembles, CNN, GRU and Attention. We perform extensive experiments for complex text classification tasks using diverse datasets for topic classification, argumentation mining and textual churn detection. Our experiments validate that ECGA is an end-to-end model that achieves or surpasses the existing state-of-the-art performance for manifold text classification tasks.

\bibliography{naaclhlt2019}

\begin{thebibliography}{20}
\expandafter\ifx\csname natexlab\endcsname\relax\def\natexlab#1{#1}\fi

\bibitem[{Amiri and Daum{\'e}~III(2015)}]{amiri2015target}
Hadi Amiri and Hal Daum{\'e}~III. 2015.
\newblock Target-dependent churn classification in microblogs.
\newblock In \emph{AAAI}, pages 2361--2367.

\bibitem[{Amiri and Daum{\'e}~III(2016)}]{amiri2016short}
Hadi Amiri and Hal Daum{\'e}~III. 2016.
\newblock Short text representation for detecting churn in microblogs.
\newblock In \emph{AAAI}, pages 2566--2572.

\bibitem[{Bojanowski et~al.(2016)Bojanowski, Grave, Joulin, and
  Mikolov}]{bojanowski2016enriching}
Piotr Bojanowski, Edouard Grave, Armand Joulin, and Tomas Mikolov. 2016.
\newblock Enriching word vectors with subword information.

\bibitem[{Breiman(2001)}]{breiman2001random}
Leo Breiman. 2001.
\newblock Random forests.
\newblock \emph{Machine learning}, 45(1):5--32.

\bibitem[{Cho et~al.(2014)Cho, van Merrienboer, Gulcehre, Bahdanau, Bougares,
  Schwenk, and Bengio}]{cho2014learning}
Kyunghyun Cho, Bart van Merrienboer, Caglar Gulcehre, Dzmitry Bahdanau, Fethi
  Bougares, Holger Schwenk, and Yoshua Bengio. 2014.
\newblock Learning phrase representations using rnn encoder--decoder for
  statistical machine translation.
\newblock In \emph{Proceedings of the 2014 Conference on Empirical Methods in
  Natural Language Processing (EMNLP)}, pages 1724--1734.

\bibitem[{Fierro et~al.(2017)Fierro, Fuentes, P{\'e}rez, and
  Quezada}]{Fierro2017}
Constanza Fierro, Claudio Fuentes, Jorge P{\'e}rez, and Mauricio Quezada. 2017.
\newblock 200k+ crowdsourced political arguments for a new chilean
  constitution.
\newblock In \emph{Proceedings of the 4th Workshop on Argument Mining}, pages
  1--10.

\bibitem[{Gridach et~al.(2017)Gridach, Haddad, and Mulki}]{gridach2017churn}
Mourad Gridach, Hatem Haddad, and Hala Mulki. 2017.
\newblock Churn identification in microblogs using convolutional neural
  networks with structured logical knowledge.
\newblock In \emph{Proceedings of the 3rd Workshop on Noisy User-generated
  Text}, pages 21--30.

\bibitem[{Howard and Ruder(2018)}]{howard2018fine}
Jeremy Howard and Sebastian Ruder. 2018.
\newblock Fine-tuned language models for text classification.

\bibitem[{Johnson and Zhang(2016)}]{johnson2016supervised}
Rie Johnson and Tong Zhang. 2016.
\newblock Supervised and semi-supervised text categorization using lstm for
  region embeddings.
\newblock In \emph{International Conference on Machine Learning}, pages
  526--534.

\bibitem[{Johnson and Zhang(2017)}]{johnson2017deep}
Rie Johnson and Tong Zhang. 2017.
\newblock Deep pyramid convolutional neural networks for text categorization.
\newblock In \emph{Proceedings of the 55th Annual Meeting of the Association
  for Computational Linguistics (Volume 1: Long Papers)}, volume~1, pages
  562--570.

\bibitem[{Joulin et~al.(2016)Joulin, Grave, Bojanowski, and
  Mikolov}]{joulin2016bag}
Armand Joulin, Edouard Grave, Piotr Bojanowski, and Tomas Mikolov. 2016.
\newblock Bag of tricks for efficient text classification.

\bibitem[{LeCun et~al.(1998)LeCun, Bottou, Bengio, and
  Haffner}]{lecun1998gradient}
Yann LeCun, L{\'e}on Bottou, Yoshua Bengio, and Patrick Haffner. 1998.
\newblock Gradient-based learning applied to document recognition.
\newblock \emph{Proceedings of the IEEE}, 86(11):2278--2324.

\bibitem[{Li et~al.(2017)Li, Gao, Wen, Du, Liu, and Wang}]{li2017joint}
Minglan Li, Yang Gao, Hui Wen, Yang Du, Haijing Liu, and Hao Wang. 2017.
\newblock Joint rnn model for argument component boundary detection.

\bibitem[{Luong et~al.(2015)Luong, Pham, and Manning}]{D15-1166}
Thang Luong, Hieu Pham, and Christopher~D. Manning. 2015.
\newblock Effective approaches to attention-based neural machine translation.
\newblock In \emph{Proceedings of the 2015 Conference on Empirical Methods in
  Natural Language Processing}, pages 1412--1421. Association for Computational
  Linguistics.

\bibitem[{Maas et~al.(2011)Maas, Daly, Pham, Huang, Ng, and
  Potts}]{maas-EtAl:2011:ACL-HLT2011}
Andrew~L. Maas, Raymond~E. Daly, Peter~T. Pham, Dan Huang, Andrew~Y. Ng, and
  Christopher Potts. 2011.
\newblock Learning word vectors for sentiment analysis.
\newblock In \emph{Proceedings of the 49th Annual Meeting of the Association
  for Computational Linguistics: Human Language Technologies}, pages 142--150,
  Portland, Oregon, USA. Association for Computational Linguistics.

\bibitem[{Socher et~al.(2013)Socher, Perelygin, Wu, Chuang, Manning, Ng, and
  Potts}]{socher2013recursive}
Richard Socher, Alex Perelygin, Jean Wu, Jason Chuang, Christopher~D Manning,
  Andrew Ng, and Christopher Potts. 2013.
\newblock Recursive deep models for semantic compositionality over a sentiment
  treebank.
\newblock In \emph{Proceedings of the 2013 conference on empirical methods in
  natural language processing}, pages 1631--1642.

\bibitem[{Srivastava et~al.(2014)Srivastava, Hinton, Krizhevsky, Sutskever, and
  Salakhutdinov}]{srivastava2014dropout}
Nitish Srivastava, Geoffrey Hinton, Alex Krizhevsky, Ilya Sutskever, and Ruslan
  Salakhutdinov. 2014.
\newblock Dropout: A simple way to prevent neural networks from overfitting.
\newblock \emph{The Journal of Machine Learning Research}, 15(1):1929--1958.

\bibitem[{Yang et~al.(2016)Yang, Yang, Dyer, He, Smola, and
  Hovy}]{yang2016hierarchical}
Zichao Yang, Diyi Yang, Chris Dyer, Xiaodong He, Alex Smola, and Eduard Hovy.
  2016.
\newblock Hierarchical attention networks for document classification.
\newblock In \emph{Proceedings of the 2016 Conference of the North American
  Chapter of the Association for Computational Linguistics: Human Language
  Technologies}, pages 1480--1489.

\bibitem[{Yin et~al.(2017)Yin, Kann, Yu, and Sch{\"u}tze}]{yin2017comparative}
Wenpeng Yin, Katharina Kann, Mo~Yu, and Hinrich Sch{\"u}tze. 2017.
\newblock Comparative study of cnn and rnn for natural language processing.

\bibitem[{Zhang et~al.(2015)Zhang, Zhao, and LeCun}]{zhang2015character}
Xiang Zhang, Junbo Zhao, and Yann LeCun. 2015.
\newblock Character-level convolutional networks for text classification.
\newblock In \emph{Advances in neural information processing systems}, pages
  649--657.

\end{thebibliography}
\bibliographystyle{acl_natbib}

\end{document}